\documentclass{article}
\usepackage{ijcai23}

\usepackage{times}
\usepackage{soul}
\usepackage{url}
\usepackage[hidelinks]{hyperref}
\usepackage[utf8]{inputenc}
\usepackage[small]{caption}
\usepackage{graphicx}
\usepackage{amsmath}
\usepackage{amsthm}
\usepackage{booktabs}
\usepackage{algorithm}
\usepackage{algorithmic}
\usepackage[switch]{lineno}
\usepackage{color}

\usepackage{amssymb}

\title{Dual Relation Knowledge Distillation for Object Detection}

\author{
Zhen-Liang Ni$^1$$^*$
\and
Fukui Yang$^2$$^*$
\and
Shengzhao Wen$^2$\and
Gang Zhang$^2$
\affiliations
$^1$ Institute of Automation, Chinese Academy of Sciences\\
$^2$ Department of Computer Vision Technology (VIS), Baidu Inc.
\emails
nizhenliang@outlook.com,
\{yangfukui, wenshengzhao, zhanggang03\}@baidu.com
}
\begin{document}

\maketitle

\begin{abstract}
    Knowledge distillation is an effective method for model compression. However, it is still a challenging topic to apply knowledge distillation to detection tasks. There are two key points resulting in poor distillation performance for detection tasks. One is the serious imbalance between foreground and background features, another one is that small object lacks enough feature representation. To solve the above issues, we propose a new distillation method named dual relation knowledge distillation (DRKD), including pixel-wise relation distillation and instance-wise relation distillation.
    The pixel-wise relation distillation embeds pixel-wise features in the graph space and applies graph convolution to capture the global pixel relation. By distilling the global pixel relation, the student detector can learn the relation between foreground and background features, and avoid the difficulty of distilling features directly for the feature imbalance issue.
    Besides, we find that instance-wise relation supplements valuable knowledge beyond independent features for small objects. Thus, the instance-wise relation distillation is designed, which calculates the similarity of different instances to obtain a relation matrix. More importantly, a relation filter module is designed to highlight valuable instance relations.
    The proposed dual relation knowledge distillation is general and can be easily applied for both one-stage and two-stage detectors. Our method achieves state-of-the-art performance, which improves Faster R-CNN based on ResNet50 from 38.4\% to 41.6\% mAP and improves RetinaNet based on ResNet50 from 37.4\% to 40.3\% mAP on COCO 2017.

\end{abstract}

\section{Introduction}
In recent years, with the development of deep learning technology, object detection has made great progress~\cite{centernet,reppoints,cornernet,fcos,faster_rcnn,cascade_rcnn}. The
detection framework can be roughly divided into two types, one-stage detector~\cite{fcos,retinanet} and two-stage detector~\cite{faster_rcnn,cascade_rcnn,cornernet,centernet}.
These deep learning methods achieve excellent performance and far surpass traditional detection methods~\cite{faster_rcnn,retinanet,centernet}. However, these deep learning methods need high computational costs, limiting their deployment on mobile devices such as robots and mobile phones. How to balance the computational cost and detection performance is still a challenging topic. Knowledge distillation is an effective method to solve the above problem~\cite{defeat,non_local_distill,fgfi,heo2019comprehensive,chen2017learning}. It adopts the form of teacher-student learning to transfer knowledge from a large model to a small model. Usually, the student model can be deployed on mobile devices directly. Since the principle of knowledge distillation is simple and effective, it is widely used in computer vision tasks, such as classification, segmentation~\cite{fgfi,defeat} and object detection.

\begin{figure}[tbp]
\centering
\includegraphics[width=\columnwidth]{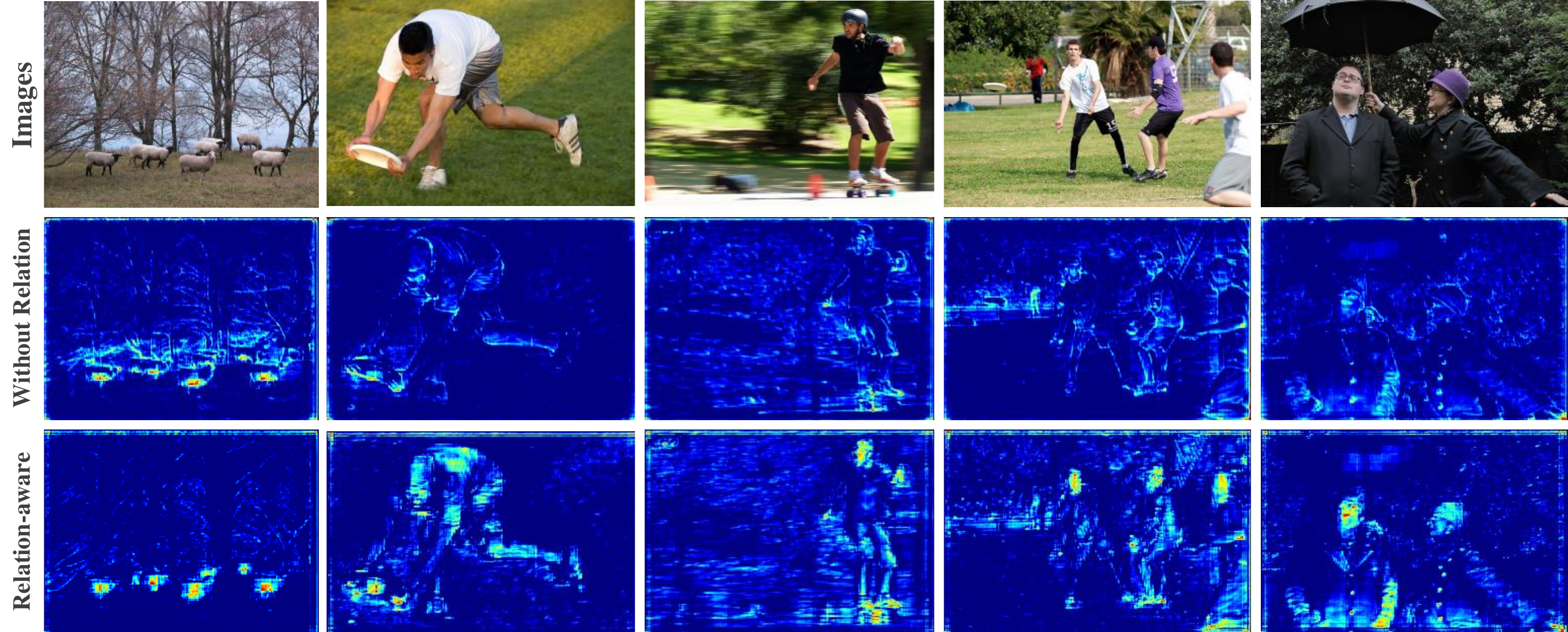}
\caption{Visualization of pixel-wise relation features. The second row and the third row show common features and pixel-wise relation features, respectively. In pixel-wise relation features, the foreground features are highlighted, proving that pixel-wise relation distillation can make the detector focus on the foreground.}
\label{heatmap}
\end{figure}

\begin{figure}[tbp]
\centering
\includegraphics[width=\columnwidth]{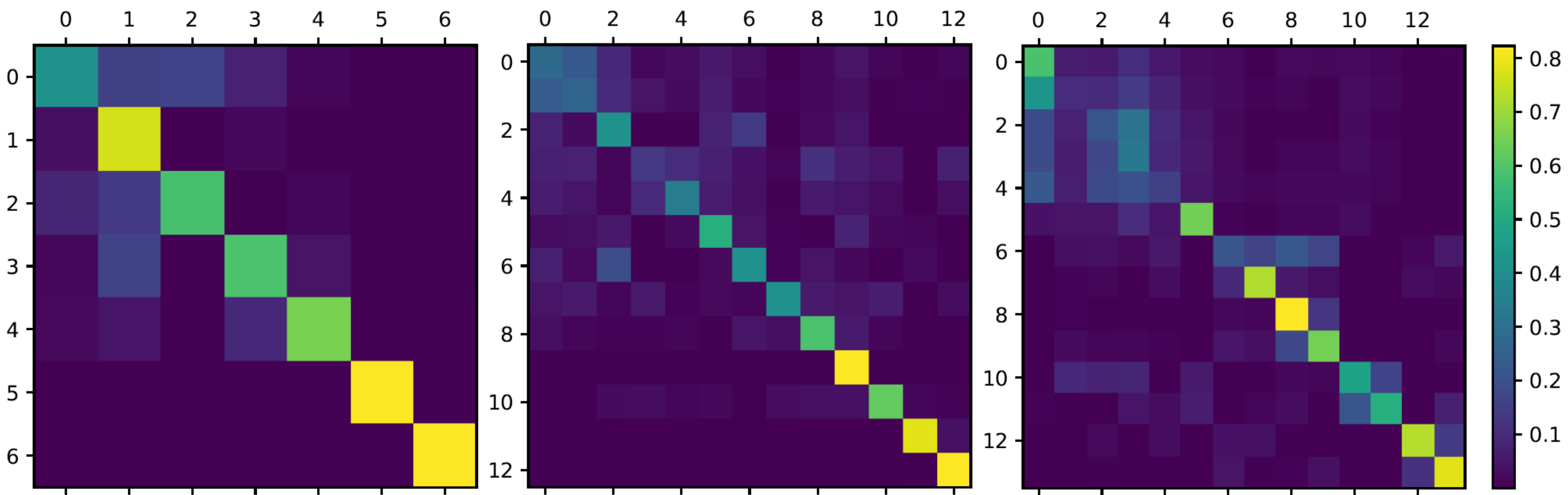}
\caption{Visualization of instance-wise relation matrix. We select 3 images that contain 7 instances, 13 instances, and 13 instances respectively. The numbers on the coordinate axis
indicate the index of instances in the same image. The index is sorted according to the size of the instance in the same image. For example, index 0 refers to the smallest instance, the
index 6 or index 12 refers to the largest instance.  We find that there are more mosaics between small instances which means that instance-wise relation supplements valuable knowledge beyond independent features for small objects. The small instance has richer relations with other size instances.}
\label{ff}
\end{figure}

However, knowledge distillation still faces many challenges in object detection~\cite{gid,non_local_distill,defeat,fgfi}. The imbalance between foreground and background is an important issue~\cite{non_local_distill,defeat}. Usually, the foreground pixels are far fewer than the background pixels. In the existing knowledge distillation methods \cite{fgfi,heo2019comprehensive}, the student model learns all pixel features with the same priority from the teacher model. So more attention will be paid to the background feature, limiting the learning of the foreground feature. Since foreground features are critical for detection, distilling all features directly results in poor performance. Some work attempts to solve this issue. NLD~\cite{non_local_distill} extracts attention features for distillation to make the detector focus on the target area. The distillation method called DeFeat~\cite{defeat} attempts to distill foreground and background features separately. These methods achieve certain results, but do not consider the relationship between instances.
Besides, it is difficult to distill effective knowledge from small instance features, resulting in poor performance of small object detection.
Current distillation methods~\cite{non_local_distill,gid} rarely take this issue into account, which limits their performance.
% However, none of them considers the relation between the foreground and the
% background, which is crucial to solve this issue.
% Previous work~\cite{rkd,relation_network,non_local_distill} has indicated that relation can transfer effective knowledge.

To address the above problems, we propose dual relation knowledge distillation (DRKD) to enable the student model to learn the relation between pixels and instances from the teacher model. We observe that pixel-wise relation is not sensitive to the imbalance between foreground and background features and can make the detector focus on the foreground, which is shown in Figure~\ref{heatmap}. Moreover, instance-wise relations can provide valuable knowledge beyond independent features, especially for small objects. As illustrated in Figure~\ref{ff}, there are 7 objects in the first image. The numbers on the coordinate axis indicate the index of instances in the same image. The index is sorted according to the size of instances in the same image. We find that there are more mosaics between small instances which means that instance-wise relation supplements valuable knowledge beyond independent features for small objects. Thus, we design two relation distillation called pixel-wise relation distillation and instance-wise relation distillation.

The pixel-wise relation distillation is proposed to make the detector focus on the learning of foreground features. We adopt graph convolution to capture global pixel relations. The graph convolution~\cite{glore} captures better feature representations than the attention module to improve model performance. First, the features from the coordinate space are embedded into the graph space. Then, the graph convolution is applied to capture relation features in graph space. Finally, the relation feature is projected back to the original coordinate space. The output feature is called the pixel-wise relation feature. By distilling the pixel-wise relation feature, the detector can pay more attention to the learning of the foreground features, as shown in Figure~\ref{heatmap}, addressing the imbalance issue.

The instance-wise relation distillation is designed to get richer representation for small instances, based on the fact that the small instance has richer relations with other size instances. First, the Embedded Gaussian function is applied to evaluate the relation of different instances, which is widely used in the attention mechanism~\cite{non_local,dan}. The similarity between different size instances is calculated to get the relation matrix. Besides, we observe that different relations have different contributions to distillation in the experiment. Thus, a relation filter module is designed to emphasize valuable relations. The filtered relation matrix is distilled to transfer instance-wise relation from the teacher detector to the student detector. Meanwhile, the cropped foreground feature is used for distillation to further improve detection accuracy in our framework. The experiment proves that the detection accuracy of small objects can be improved by instance-wise relation distillation.
The contributions of this work can be concluded as follows:
\begin{itemize}
\item We propose pixel-wise relation distillation based on graph convolution. The graph convolution can capture global context more efficiently than the attention mechanism, achieving a better distillation effect.
%The global pixel relations are distilled to make the detector focus on the foreground, addressing the feature imbalance issue.
%This method captures global pixel relations to make the detector focus on the foreground, addressing the feature imbalance issue.
\item We propose instance-wise relation distillation to get richer representation for small instances. The experiment proves that the detection accuracy of small objects can be improved by instance-wise relation distillation.
% We observe that small instance has richer relations with other size instances.
\item Our dual relation knowledge distillation achieves state-of-the-art performance, which improves Faster R-CNN based on ResNet50 from 38.4\% to 41.6\% mAP and improves RetinaNet based on ResNet50 from 37.4\% to 40.3\% mAP on COCO2017. The proposed dual relation knowledge distillation is general and can be easily applied for both one-stage and two-stage detectors.
\end{itemize}

\begin{figure*}[htbp]
  \centering
  \includegraphics[width=0.96\textwidth]{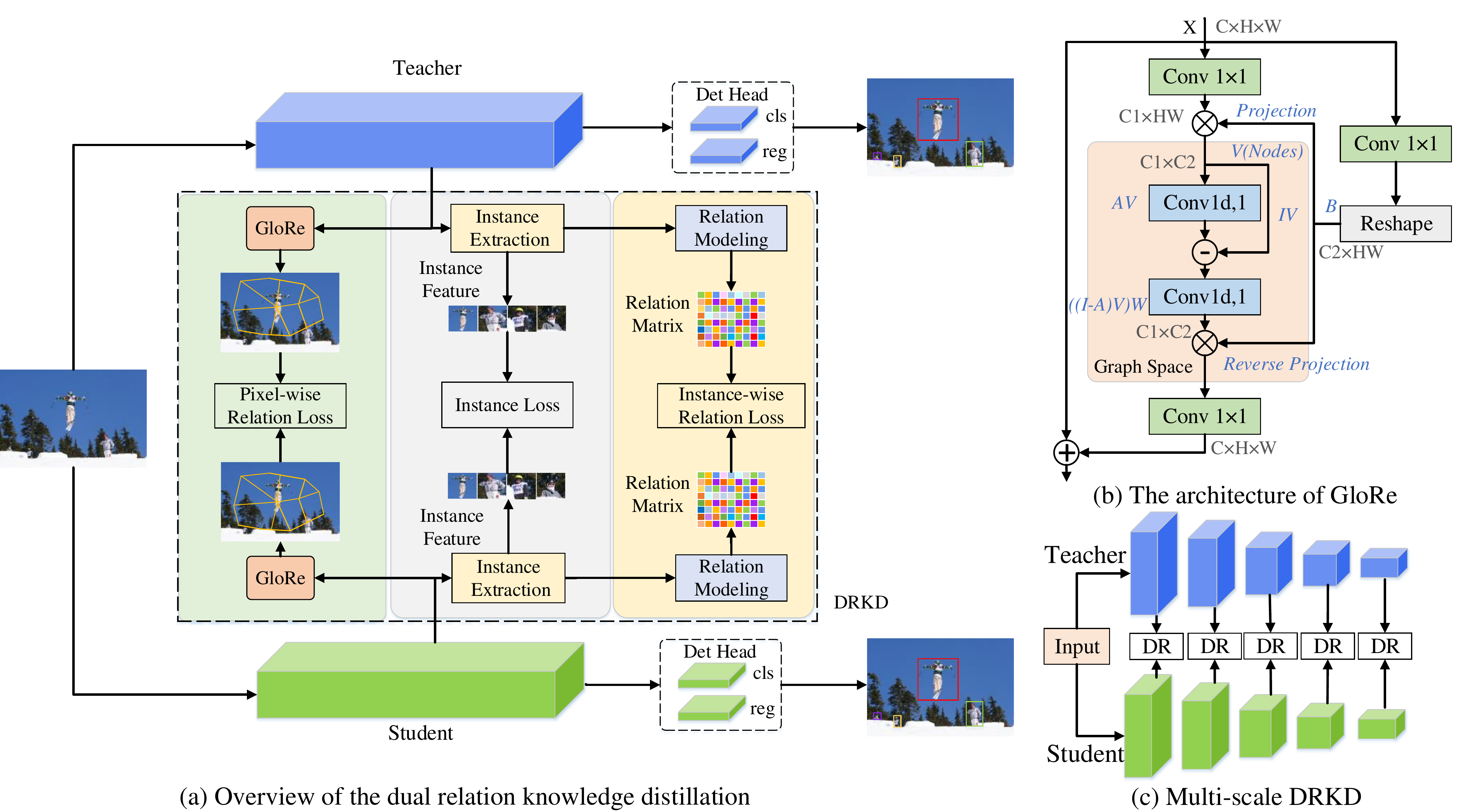}
  \caption{(a) The overview of the proposed dual relation knowledge distillation (DRKD). DRKD includes pixel-wise relation distillation, instance-wise relation distillation and instance distillation. The pixel-wise relation is captured by the graph convolution module GloRe. The instance-wise relation is modeled by calculating the similarity between instance features;
  (b) shows the architecture of GloRe used in pixel-wise relation distillation;  (c) shows multi-scale DRKD.}
  \label{framework}
\end{figure*}
\section{Related Work}
\subsection{Object Detection}
Recently, object detection\cite{centernet,reppoints,cornernet} is widely studied. The researchers first propose two-stage detectors~\cite{faster_rcnn,cascade_rcnn}, generating candidate regions for classification and location. Two-stage detectors usually have higher accuracy but lower speed. The representative methods are Faster R-CNN~\cite{faster_rcnn} and Cascade R-CNN~\cite{cascade_rcnn}. To solve the problem of two-stage detectors, one-stage detectors are proposed furtherly such as SSD~\cite{ssd}, YOLOv3~\cite{yolov3}, RetinaNet~\cite{retinanet}, and Fcos~\cite{fcos} which do not require region proposals and can directly obtain the final detection results. Thus, one-stage detectors have higher speed but poor performance. To balance speed and performance, model compression is used in detection tasks, such as pruning~\cite{han2015learning,lin2020hrank}, quantification~\cite{zhang2021diversifying,gong2019differentiable}, and distillation~\cite{non_local_distill,gid}. Among them, distillation is the most widely used. However, the distillation still faces many challenges in the detection task. The main challenge is to solve the imbalance issue and improve the small object detection performance. Thus, we propose dual relation distillation.

\subsection{Knowledge Distillation}
Knowledge distillation is a common method of model compression. It can transfer the knowledge from the teacher detector to improve the performance of the student detector. However, it is still a challenging topic to apply knowledge distillation to object detection directly due to some problems such as the imbalance of foreground and background, and the small object detection problem. Some researchers attempt to apply distillation to object detection. Sun et al~\cite{tar} not only distilled hidden features but also distilled classification output and bounding box regression output to improve detection accuracy. FGFI~\cite{fgfi} proposed a fine-grained feature imitation for anchor-based detectors. It calculated the regions of the near object by the intersection between target boxes and anchors. DeFeat~\cite{defeat} decoupled features into foreground and background and distilled them separately to get better distillation performance. General instance distillation (GID)~\cite{gid} proposed the general instance selection module and modeled the relational knowledge between instances for distillation. Non-local distillation~\cite{non_local_distill} applied non-local modules to capture global pixel relation and enables the student to learn the pixel relation from the teacher. Different from them, we not only distilled the pixel-wise relations and instance-wise relations but also designed more effective modules to extract them.

\section{Method}
The dual relation knowledge distillation (DRKD) is proposed to enable the student detector to learn the pixel-wise and instance-wise relations from the teacher detector, which is shown in Figure~\ref{framework}.

\subsection{Pixel-wise Relation Distillation}
Pixel-wise relation distillation helps student detectors learn the relation between foreground and background features, addressing the feature imbalance issue. The graph convolution module, named GloRe~\cite{glore}, is adopted to capture the global pixel relation. It can capture global context more efficiently than the attention mechanism~\cite{glore}, achieving a better distillation effect.
% By distilling the relation-aware features generated by the graph convolution, the positioning ability of the student detector can be improved.

Specifically, we extract multi-scale features from the backbone of teachers and students respectively, and feed them to different GloRe modules for capturing the global pixel relation.
After that, the pixel-wise relation features are distilled to transfer the global relation from the teacher to the student. The distillation loss is shown in Equ. (1). Besides, to minimize
the feature difference between the student and teacher models, an adaptive convolution is added on the side of the student model.
% $L_2$ norm loss is used for distillation, which is shown in Equ. (1).
\begin{equation}
{L_{PR}} = \frac{1}{k}\sum\limits_{i = 1}^k {\left\| {\phi ({t_i}) -  f(\phi ({s_i}))} \right\|_2}
\end{equation}
where $k$ is the number of features. $t_i$ and $s_i$ refer to the feature from the teacher and student respectively. $\phi$ represents the GloRe module. $f$ represents adaptive
convolution.

As shown in Figure~\ref{framework} (b), GloRe contains three parts: graph embedding, graph convolution, and reprojection. The coordinate feature is first projected to a low-dimensional
graph feature space.
For the input feature $X \in {R^{C \times W \times H}}$, we first project and transform it to $\overline X \in {R^{C1 \times HW}}$ by linear layer. After that, the graph node features $V\in {R^{C1 \times C2}}$ can be obtained by projecting $\overline X$. The projection matrix is $B  \in{R^{C2 \times HW}}$.
The projection method is a linear combination of original features, which is shown in Equ. (2). The graph node features can aggregate information from multiple regions.
\begin{equation}
%v_i=b_{i}\overline X=\sum\limits_{\forall j}{b_{ij} x_j}
V = \overline XB^{T}
\end{equation}
where $B \in {R^{C2 \times HW}}$ is learnable projection matrix. $V \in {R^{C1 \times C2}}$ is the graph node features.

Based on graph node features, a graph convolution is used to capture the relation between nodes, which is defined by Equ. (3). $A$ denotes the adjacency matrix, which is initialized randomly and updated with training. In the training process, the adjacency matrix learns the weights of the edges between nodes. The weight of edges reflects the relation of nodes. Based on the adjacency matrix and the state update matrix, the node features are updated to obtain the relation-aware features.
\begin{equation}
Z = \left( {\left( {I - A} \right)V} \right)W
\end{equation}
where $I \in {R^{C1 \times C1}}$ is an identity matrix. $A \in{ R^{C1 \times C1}}$ represents nodes adjacency matrix. $V \in {R^{C1 \times C2}}$ is the graph node features. $W \in {R^{C2 \times C2}}$ represents the state update matrix. $Z \in {R^{C1 \times C2}}$ represents the relation-aware features in graph space.

Finally, the relation-aware features are projected back to the coordinate feature space, as shown in Equ. (4).
\begin{equation}
F = ZB
\end{equation}
where $F \in {R^{C1 \times HW}}$ is pixel-wise relation feature, $B \in {R^{C2 \times HW}}$ is learnable projection matrix, the same as Equ. (2). $Z \in {R^{C1 \times C2}}$ represents the relation-aware features in graph space.

\subsection{Instance-wise Relation Distillation}
The instance-wise relation distillation is designed to get richer representation for small instances, based on the fact that the small instance has richer relations with other size instances. The Embedded Gaussian function is applied to model the similarity of instance features. Moreover, we design a relation filter module to emphasize valuable relations. The instance-wise relation module is shown as Figure~\ref{object_relation}.

\subsubsection{Instance Feature Extraction}
To capture instance-wise relation, we need to extract the instance features. The ground truth of object coordinates is used to extract instance features from the feature map, according to the ratio of the input and the feature map. The extracted instance features are resized to the same size, which is shown in Equ. (5).
\begin{equation}
\widehat x = \xi (x,c,o)
\end{equation}
where $\xi$ represents ROI Align. $x$ is input feature map. $c$ represents coordinates of ground truth. $o$ indicates the size of the output feature.

\subsubsection{Instance-wise Relation Module and Distillation}
\begin{figure}[tbp]
  \centering
  \includegraphics[width=235pt]{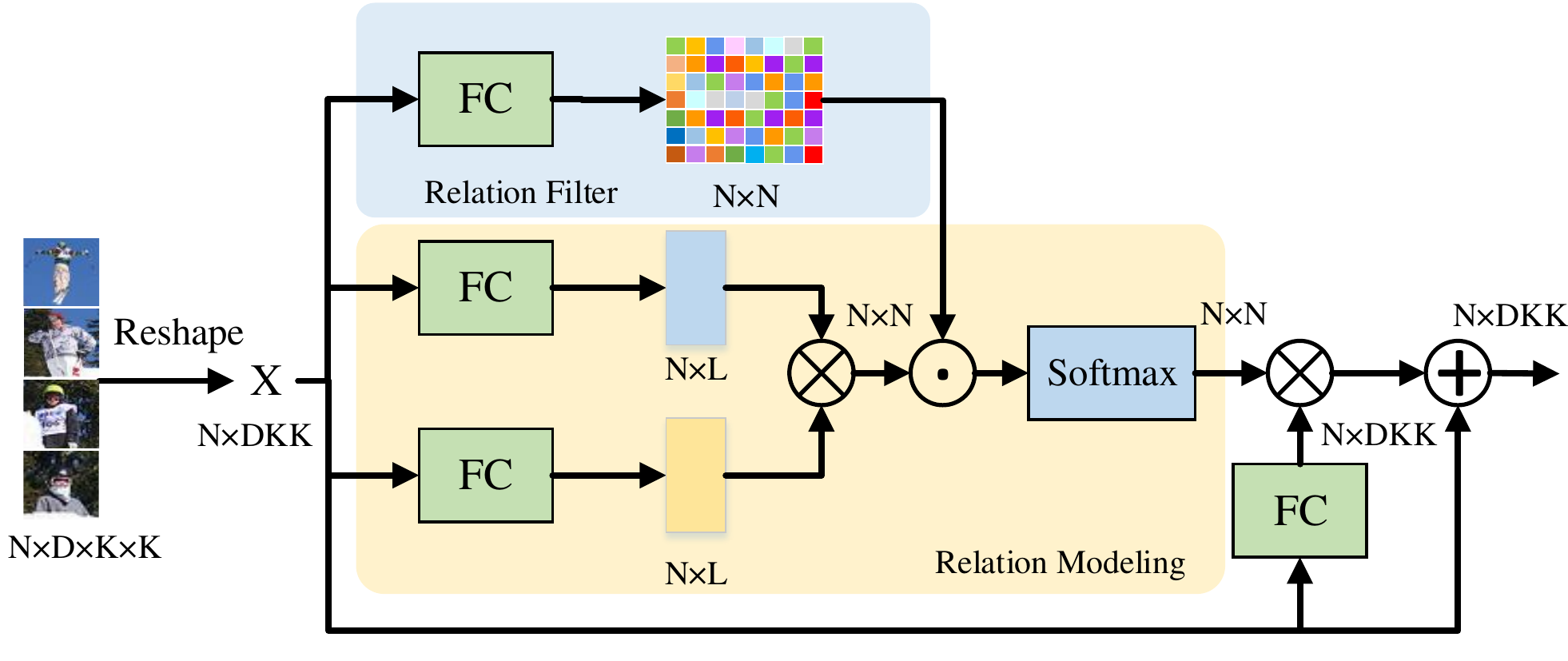}
  \caption{The architecture of instance-wise relation module. The similarity of different instances is calculated to obtain the relation matrix. A relation filter module is designed to emphasize
  valuable relation. }
  \label{object_relation}
\end{figure}
As shown in Figure~\ref{object_relation}, the input feature is $X \in {R^{N \times DKK}}$. N refers to the number of instances in the same image, D refers to the channel of the instance feature, K refers to the size of the instance feature map. The instance-wise relation module can be represented by Equ. (6).
\begin{equation}
   \psi \left( {{s_i},{s_j}} \right) = \frac{1}{\tau }{e^{{w_{ij}}{g_1}({s_i}){g_2}({s_j})}},\tau  = \sum\nolimits_{\forall i} {{e^{{w_{ij}}{g_1}({s_i}){g_2}({s_j})}}}
\end{equation}
where $s_i$ and $s_j$ refer to the instance feature. $g_1$  and $g_2$ refer to fully connected layer. $w_{ij}$ is the weight from relation matrix  $W \in {R^{N \times N}}$ as shown in relation filter of Figure~\ref{object_relation}. $\psi \left( {{s_i},{s_j}} \right)$ refers to instance-wise relation feature between instance $s_i$ and $s_j$. We define instance-wise relation distillation function as   Equ. (7).

\begin{equation}
    {L_{IR}} = \sum\limits_{(i,j) \in {{\rm N}^2}} {{{\left\| {\psi \left( {{t_i},{t_j}} \right) - f\left(\psi \left( {{s_i},{s_j}} \right)\right)} \right\|}_2}}
\end{equation}
where $s_i$ and $s_j$ refer to the instance feature from student model and $t_i$ and $t_j$ refer to the instance feature from teacher model. $\psi \left( {{s_i},{s_j}} \right)$ refers to instance-wise relation feature between instance $s_i$ and $s_j$. $\psi \left( {{t_i},{t_j}} \right)$ refers to instance-wise relation feature between instance $t_i$ and $t_j$. $f$ represents adaptive convolution.
\begin{table*}[tbp]
  \centering
  \resizebox{0.9\textwidth}{!}{
    \begin{tabular}{cccccccccccc}
    \hline
    \hline
   Epochs&Framewark &Model & PR & IR & INS &$mAP$ & $AP_{50}$  & $AP_{75}$  & $AP_{S}$  & $AP_{M}$  & $AP_{L}$ \\
    \hline
    12&One-stage&Teacher(RX101) &- &-  & - & 41.0  & 60.9  & 44.0  & 23.9  & 45.2  & 54.0 \\
    \hline
    12&One-stage&Student(R50)  &$\times$ &$\times$  & $\times$  & 36.5  & 55.4  & 39.1  & 20.4  & 40.3  & 48.1 \\
    12&One-stage&Student(R50)  &\checkmark &$\times$  & $\times$  & 38.0  & 56.8  & 40.7  & 22.0    & 41.8  & 50.9 \\
    12&One-stage&Student(R50)  &$\times$ &\checkmark  & $\times$  & 36.9  & 55.9  & 39.4  & 21.2  & 40.7  & 47.8 \\
    12&One-stage&Student(R50)  &\checkmark &\checkmark  &$\times$  & 38.4  & 57.6  & 40.9  &  \textbf{22.6}  & 42.1  &  \textbf{51.0} \\
    12&One-stage&Student(R50)  &\checkmark &\checkmark &\checkmark   &  \textbf{38.5}  &  \textbf{57.6}  &  \textbf{41.1}  & 21.6  & \textbf{42.6}  & 50.8 \\
    \hline
    24&One-stage&Student(R50) &$\times$ &$\times$  & $\times$& 37.4  & 56.7  & 39.6  & 20.0 & 40.7  & 49.7 \\
    24&One-stage&Student(R50) &\checkmark &$\times$  & $\times$& 39.9  & 59.2  & 42.8  & 22.3  & 43.9  &  \textbf{53.7} \\
    24&One-stage&Student(R50) &$\times$ &\checkmark  & $\times$& 38.3  & 57.5  & 40.8  & 21.2  & 41.9  & 50.6 \\
    24&One-stage&Student(R50) &\checkmark &\checkmark  &$\times$& 40.1  & 59.4  & 42.9  &  \textbf{23.8}  & 43.9  & 53.6 \\
    24&One-stage&Student(R50) &\checkmark &\checkmark &\checkmark &  \textbf{40.3}  &  \textbf{59.7}  &  \textbf{42.9}  & 23.4  &  \textbf{44.2}  & 53.4\\
    \hline
    12&Two-stage&Teacher(RX101) &- &-  & - & 42.1  & 63.0    & 46.3  & 24.8  & 46.2  & 55.3 \\
    \hline
    12&Two-stage&Student(R50) &$\times$ &$\times$  & $\times$ & 37.4  & 58.1  & 40.4  & 21.2  & 41.0    & 48.1 \\
    12&Two-stage&Student(R50) &\checkmark &$\times$  & $\times$& 39.4  & 60.2  & 43.1  & 22.3  & 43.4  & 51.8 \\
    12&Two-stage&Student(R50) &$\times$ &\checkmark  & $\times$& 38.2  & 59.1  & 41.6  & 22.3  & 41.8  & 49.4 \\
    12&Two-stage&Student(R50) &\checkmark &\checkmark  &$\times$& 39.6  &  \textbf{60.6}  & 42.9  & 22.6  &  \textbf{43.7}  & 51.7 \\
    12&Two-stage&Student(R50) &\checkmark &\checkmark &\checkmark& \textbf{39.7}  & 60.5  &  \textbf{43.1}  & \textbf{23.5}  & 43.4  &  \textbf{52.3} \\
    \hline
    \hline
    \end{tabular}
    }
   \caption{Distillation results of both one-stage detector RetinaNet and two-stage detector Faster RCNN on COCO2017. PR refers to Pixel-wise Relation distillation loss, IR refers to Instance-wise Relation distillation loss, INS refers to Instance distillation loss. RX101 refers to ResNeXt101, R50 refers to ResNet50. One-stage framework refers to RetinaNet. Two-stage framework refers to Faster RCNN.}
  \label{ablation}
\end{table*}

Furthermore, instance features also help address the imbalance issue between foreground and background~\cite{defeat}. Instance distillation can make the detector pay attention to the learning of foreground features and accelerate the convergence of the student detector~\cite{gid}. Therefore, we directly distill the instance features to further improve the distillation performance. $L_2$ norm loss is adopted for distillation, which is shown in Equ. (8). The adaptive convolution is applied on the side of the student detector to minimize the feature difference between the student and teacher.
\begin{equation}
{L_{INS}} = \frac{1}{n}\sum\limits_{i = 1}^n {{{\left\| {{t_i} - f({s_i})} \right\|}_2}}
\end{equation}
where $n$ refers to the number of foreground features. $t_{i}$ and $s_{i}$ refer to the feature from the teacher model and student model, respectively. $f$ represents adaptive convolution.

\begin{algorithm}[tbp]
	%\textsl{}\setstretch{1.8}
	\renewcommand{\algorithmicrequire}{\textbf{Input:}}
	\renewcommand{\algorithmicensure}{\textbf{Output:}}
	\caption{Dual Relation Knowledge Distillation}
	\label{alg1}
	\begin{algorithmic}[1]
		\STATE Initialize ${L_{PR}}$, ${L_{IR}}$, and ${L_{INS}}$ to $0$.
		\STATE Calculate the loss ${L_{PR}}$ based on Equation (1).
		\STATE Extract teacher instances ${t}$ based on Equation (5).
		\STATE Extract student instances ${s}$ based on Equation (5).
		\STATE for ${i}$ in 0:length(${t}$)
		\STATE$\qquad$for ${j}$ in 0:length(${t}$)
		\STATE$\qquad$$\qquad$Calculate $R_t=\psi \left( {t_{i}, t_{j}}\right)$ based on Equation (6).
		\STATE$\qquad$$\qquad$Calculate $R_s=\psi \left( {s_{i}, s_{j}}\right)$ based on Equation (6).
		\STATE Calculate the loss ${L_{IR}}$ based on Equation (7).
		\STATE Calculate the loss ${L_{INS}}$ based on Equation (8).
		\STATE Calculate overall loss $L$ based on Equation (9).
	\end{algorithmic}
\end{algorithm}
\subsection{Overall Loss Function}
According to the above analysis, the overall loss function contains four parts. $L_{det}$ is the task loss used to train the detection model. ${L_{PR}}$ is the pixel-wise relation distillation loss. ${L_{IR}}$ is the instance-wise relation distillation loss. ${L_{INS}}$ is the instance distillation loss.  The overall loss function is shown in Equ. (9)
\begin{equation}
    L = {L_{\det }} + {\lambda _1}{L_{PR}} + {\lambda _2}{L_{IR}} + {\lambda _3}{L_{INS}}
\end{equation}
where ${\lambda _1},{\lambda _2},{\lambda _3}$ are hyperparameters used to balance different losses. The pseudo code is shown in Algorithm 1.

\section{Experimental and Results}
\subsection{Implementation Details}
COCO2017~\cite{coco} is used to evaluate our method, which is a challenging dataset in object detection. It contains 120k images and 80 object classes. All experiments are performed on 8 Tesla P40 GPUs. The optimizer we use is SGD. The batch size is set to 16. The initial learning rate is 0.02. The momentum is set to 0.9 and the weight decay is 0.0001. Unless specified, the ablation experiment usually adopts 1$\times$ learning schedule and the comparison experiment with other methods adopts 2$\times$ learning schedule.
We consider Average Precision as evaluation metric, i.e., $mAP$, $AP_{50}$, $AP_{75}$, $AP_S$, $AP_M$ and $AP_L$.

\subsection{Ablation Study}
To verify the effectiveness of the proposed distillation method, a series of ablation experiments are performed, which is shown in Table~\ref{ablation}. Our method is evaluated on both the one-stage detector and the two-stage detector.

For the one-stage detector, RetinaNet based on ResNeXt101~\cite{resnext} is chosen as the teacher detector and RetinaNet based on ResNet50~\cite{resnet} is selected as the student detector. Under the training strategy of 24 epochs, the student detector achieves a 37.4\% mAP. Applying pixel-wise relation distillation brings an increase of 2.5\% mAP. Adopting instance-wise relation distillation achieves an increase of 0.9\% mAP. Compared to the student detector without distillation, employing the proposed DRKD increases mAP by 2.9\%. For small object detection, applying pixel-wise relation distillation and instance-wise relation distillation simultaneously achieves an increase of 3.8\% $AP_S$ from 20.0\% $AP_S$ to 23.8\% $AP_S$.

For the two-stage detector, Faster RCNN based on ResNeXt101 is chosen as the teacher detector and Faster RCNN based on ResNet50 is selected as the student detector. Under the training strategy of 12 epochs, the student detector achieves a 37.4\% mAP. Applying pixel-wise relation distillation brings an increase of 2.0\% mAP. Adopting instance-wise relation distillation achieves an increase of 0.8\% mAP. Compared to the student detector without distillation, the application of our DRKD increases mAP by 2.3\%. For small object detection, applying DRKD achieves an increase of 2.3\% $AP_S$ from 21.2\% $AP_S$ to 23.5\% $AP_S$. Besides, experiments prove that our pixel relation modeling method based on graph convolution is better than non-local, which is shown in Table~\ref{pixel_ablation}. The above ablation experiments prove that our method is effective on both one-stage and two-stage detectors.

\begin{table}[tbp]
  \centering
    \begin{tabular}{ccc}
    \hline
    \hline
    Model & Distillation & mAP \\
    \hline
    RX101-RetinaNet & Teacher & 41.0 \\
    R50-RetinaNet & Student & 36.5 \\
    RX101-R50-RetinaNet & Non-Local & 37.6 \\
    RX101-R50-RetinaNet & Ours & \textbf{38.0} \\
    \hline
    \hline
    \end{tabular}
  \caption{Comparison of different pixel relation modeling methods. Our pixel relation modeling module has better distillation performance than non-local. }
  \label{pixel_ablation}
\end{table}
Besides, we analyze the mean AP of objects with different sizes to determine the effects of different distillation methods. As shown in Table~\ref{ablation}, pixel-wise relation distillation brings growth in $AP_S$, $AP_M$, and $AP_L$, indicating that it can improve the detection accuracy of objects with different sizes. This result verifies that the pixel-wise relation distillation helps to address the feature imbalance issue. Moreover, based on pixel-wise relation distillation, employing instance-wise relation distillation in 24 epochs can further increase $AP_S$ of RetinaNet by 1.5\% $AP_S$ from 22.3\% $AP_S$ to 23.8\% $AP_S$. For small object detection, applying pixel-wise relation distillation and instance-wise relation distillation simultaneously achieves an increase of 3.8\% $AP_S$ from 20.0\% $AP_S$ to 23.8\% $AP_S$. Our experiment proves that the detection accuracy of small objects can be improved significantly when using both two relation distillations.

\begin{table}[tbp]
  \centering
    \begin{tabular}{cccc}
    \hline
    \hline
    Model & Distillation & Size  & mAP \\
    \hline
    R50-RetinaNet & -     & -     & 36.5 \\
    RX101-R50-RetinaNet & IR & $1 \times 1$ &  36.6\\
    RX101-R50-RetinaNet & IR & $3 \times 3$ & 36.9 \\
    RX101-R50-RetinaNet & IR & $5 \times 5$   &  36.6\\
    RX101-R50-RetinaNet & PR+IR+INS& $1 \times 1$  &  37.9\\
    RX101-R50-RetinaNet & PR+IR+INS& $3 \times 3$   & \textbf{38.5} \\
    RX101-R50-RetinaNet & PR+IR+INS& $5 \times 5$   &  38.3\\
    \hline
    \hline
    \end{tabular}
  \caption{Distillation performance comparison of different instance feature size settings. RX101-R50-RetinaNet indicates that the teacher detector is ResNeXt101 based RetinaNet, while the student detector is ResNet50 based RetinaNet.PR refers to Pixel-wise Relation distillation loss, IR refers to Instance-wise Relation distillation loss, INS refers to Instance distillation loss.}
  \label{size}
\end{table}
\subsection{Instance Feature Size Selection}
Since the instance-wise relation is modeled based on instance features, the size of instance features may affect the performance of distillation. Therefore, experiments are performed to
select the appropriate foreground feature size, which is shown in Table~\ref{size}. Three different feature sizes, including $1\times1$, $3\times3$, and $5\times5$, are set up in the experiment. When only using instance-wise relation distillation, the $3\times3$ feature size achieves the best result 36.9\% mAP, which exceeds baseline by 0.4\% mAP. When using three distillation methods, the $3\times3$ feature size also gets the best result. Therefore, the $3\times3$ size is selected in all experiments.

\subsection{Relation Filter Experiment}
\begin{table}[tbp]
  \centering
    \begin{tabular}{cccc}
    \hline
    \hline
    Model & Distillation & Filter & mAP\\
    \hline
    RX101-RetinaNet & Teacher & -     & 41.0 \\
    R50-RetinaNet & Student & -     & 36.5 \\
    RX101-R50-RetinaNet & IR   & $\times$     & 36.6 \\
    RX101-R50-RetinaNet & IR   & \checkmark     & 36.9 \\
    RX101-R50-RetinaNet & PR+IR+INS & $\times$    & 38.3 \\
    RX101-R50-RetinaNet & PR+IR+INS & \checkmark     & \textbf{38.5} \\
    \hline
    \hline
    \end{tabular}%
  \caption{Validity verification of relation filtering module in instance-wise relation distillation.  PR refers to Pixel-wise Relation distillation loss, IR refers to Instance-wise Relation distillation loss, INS refers to Instance distillation loss.}
  \label{filter}%
\end{table}%

\begin{table}[tbp]
  \centering
    \begin{tabular}{c|c|c|c|c|c}
    \hline
    \hline
    $\lambda _1$    & mAP   & $\lambda _2$    & mAP   & $\lambda _3$    & mAP \\
    \hline
    0.006 & 37.9  & 0.001 & 36.7  & 0.005 & 38.4 \\
    0.004 & 38.0  & 0.002 & 36.8  & 0.006 & 38.5 \\
    0.002 & 37.7  & 0.004 & 36.9  & 0.008 & 38.4 \\
    0.001 & 37.2  & 0.006 & 36.8  & 0.01  & 38.3 \\
    \hline
    \hline
    \end{tabular}%
  \caption{Hyper-parameter sensitivity study of $\lambda _1$, $\lambda _2$, and $\lambda _3$, with RetinaNet on COCO 2017.}
  \label{weight_choose}%
\end{table}%

\begin{table*}[tbp]
  \centering
  \resizebox{0.95\textwidth}{!}{
    \begin{tabular}[]{ccccccccc}
    \hline
    \hline
    Framewark &Method & Backbone &mAP    & $AP_{50}$  & $AP_{75}$  & $AP_{S}$   & $AP_{M}$   & $AP_{L }$ \\
    \hline
    Cascade Mask RCNN &Teacher & RX101 & 47.3  & 66.3  & 51.7  & 28.2  & 51.7  & 62.7  \\
    Faster RCNN &Baseline  & R50 & 38.4  & 59.0    & 42.0    & 21.5  & 42.1  & 50.3  \\
    Faster RCNN  &EDKD~\cite{chen2017learning}  & R50 & 38.7  & 59.0    & 42.1  & 22.0    & 41.9  & 51.0  \\
    Faster RCNN  &FGFI~\cite{fgfi}  & R50 & 39.1  & 59.8  & 42.8  & 22.2  & 42.9  & 51.1 \\
    Faster RCNN  &Overhaul~\cite{heo2019comprehensive}   & R50 & 38.9  & 60.1  & 42.6  & 21.8  & 42.7  & 50.7  \\
    Faster RCNN  &NLD\textcolor{blue}{(ICLR-21)}~\cite{non_local_distill}   & R50 & 41.5  & 62.2  & 45.1  & 23.5  & 45.0    & 55.3 \\
    Faster RCNN  &GKD\textcolor{blue}{(ECCV-22)}~\cite{Tang2022DistillingOD}   & R50 & 41.5  & 61.9  & 45.1  & 23.5  & 45.1    & \textbf{55.4} \\
    Faster RCNN  &DRKD (Ours)  & R50 & \textbf{41.6}  & \textbf{62.4}  & \textbf{45.3}  & \textbf{24.2}  & \textbf{45.3}  & 55.3 \\
    \hline
    Cascade Mask RCNN &Teacher & RX101 & 47.3  & 66.3  & 51.7  & 28.2  & 51.7  & 62.7  \\
    Grid RCNN &Baseline & R50 & 40.4  & 58.4  & 43.6  & 22.8  & 43.9  & 53.3 \\
    Grid RCNN & NLD\textcolor{blue}{(ICLR-21)}~\cite{non_local_distill}  & R50 & 42.6  & 61.1  & 46.1  & 24.2  & \textbf{46.6}  & 55.8  \\
    Grid RCNN &DRKD (Ours)  & R50 & \textbf{43.0} & \textbf{61.8}  & \textbf{46.5}  & \textbf{25.1} & 46.5  & \textbf{56.3}  \\
    \hline
    Cascade Mask RCNN &Teacher & RX101 & 47.3  & 66.3  & 51.7  & 28.2  & 51.7  & 62.7 \\
    Dynamic RCNN &Baseline & R50 & 39.8  & 58.3  & 43.2  & 23.0    & 42.8  & 52.4   \\
    Dynamic RCNN &NLD\textcolor{blue}{(ICLR-21)}~\cite{non_local_distill}   & R50 & 42.8  & 61.2  & 47.0    & 23.9  & 46.2  & \textbf{57.7}  \\
    Dynamic RCNN &DRKD (Ours)  & R50 & \textbf{43.0} & \textbf{61.6} & \textbf{47.3}  &\textbf{24.1}  &\textbf{46.3}  & 57.6  \\
    \hline
    RetinaNet &Teacher & RX101 & 41.0  & 60.9  & 44.0    & 23.9  & 45.2  & 54.0  \\
    RetinaNet &Baseline & R50 & 37.4  & 56.7  & 39.6  & 20.0    & 40.7  & 49.7  \\
    RetinaNet &Overhaul~\cite{heo2019comprehensive}   & R50 & 37.8  & 58.3  & 41.1  & 21.6  & 41.2  & 48.3  \\
    RetinaNet &NLD\textcolor{blue}{(ICLR-21)}~\cite{non_local_distill}   & R50 & 39.6  & 58.8  & 42.1  & 22.7  & 43.3  & 52.5  \\
    RetinaNet & FRS\textcolor{blue}{(NIPS-21)}~\cite{FRS}   & R50 & 40.1  & 59.5  & 42.5  & 21.9  & 43.7  & \textbf{54.3}  \\
    RetinaNet &DRKD (Ours)  & R50 & \textbf{40.3}  & \textbf{59.7}  & \textbf{42.9}  & \textbf{23.4}  & \textbf{44.2}  & 53.4  \\
    \hline
    RetinaNet &Teacher & RX101 & 41.0  & 60.9  & 44.0    & 23.9  & 45.2  & 54.0  \\
    RepPoints &Baseline & R50 & 38.6  & 59.6  & 41.6  & 22.5  & 42.2  & 50.4  \\
    RepPoints &NLD\textcolor{blue}{(ICLR-21)}~\cite{non_local_distill}   & R50 & 40.6  & 61.7  & 43.8  & 23.4  & 44.6  & 53.0  \\
    RepPoints &FGD\textcolor{blue}{(CVPR-22)}~\cite{fgd}   & R50 & 41.3  & -  & -  & -  & 45.2  & 54.0\\
    RepPoints &DRKD (Ours)  & R50 & \textbf{41.7}  & \textbf{62.5}  & \textbf{44.9}  &\textbf{24.3}  &\textbf{45.6}  &\textbf{55.0}  \\
    \hline
    \hline
    \end{tabular}%
    }
  \caption{Comparison between our methods and other state-of-the-art distillation methods. RX101 refers to ResNeXt101, R50 refers to ResNet50.}
  \label{sota}%
\end{table*}%
The relation filter module is set to highlight important instance relations. Related experiments are performed to verify its effectiveness, which is shown in Table~\ref{filter}. When only
using instance-wise relation distillation, applying relation filter results in a 0.3\% mAP increase. Based on our DRKD without relation filter, employing the relation filter brings a 0.2\% mAP
increase. The above results prove the effectiveness of the relation filter module.
%\begin{figure}[tbp]
%  \centering
%  \includegraphics[width=\columnwidth]{image/weight_chosse1.pdf}
%  \caption{Hyper-parameter sensitivity study of $\lambda _1$, $\lambda _2$, and $\lambda _3$, with RetinaNet on COCO 2017.}
%  \label{weight_choose}
%\end{figure}

\subsection{Hyperparameter Selection}
To obtain the best distillation performance, we analyze the sensitivity of hyperparameters. A series of experiments are set up to determine the value of $\lambda _1$, $\lambda _2$, and
$\lambda _3$ in Equ. (9). RetinaNet is chosen to select parameters for the one-stage detector. As shown in Table~\ref{weight_choose}, when $\lambda _1$ is 0.004, pixel-wise relation distillation gets the best performance. When $\lambda _2$ is 0.004, instance-wise relation distillation gets the best performance. When $\lambda _3$ is 0.006, instance distillation achieves the best
performance. For two-stage detectors,  $\lambda _1$ and $\lambda _3$ are multiplied by 0.25 and 0.3 respectively on the basis of the above values to obtain the best results.

\subsection{Comparison with State-of-the-art Methods}
To verify the excellent performance of the proposed method, our DRKD is compared with other state-of-the-art distillation methods on COCO2017. The experimental results are shown in
Table~\ref{sota}. Since our DRKD can be applied to both one-stage and two-stage detectors, we verify it on these two types of detectors. The experimental results show that our DRKD exceeds the current most advanced methods, achieving the state-of-the-art performance.

For the two-stage detector, the Cascaded Mask RCNN with ResNeXt 101 is selected as the teacher detector. Faster RCNN~\cite{faster_rcnn}, Dynamic RCNN~\cite{dynamic_rcnn}, and Grid RCNN~\cite{grid_rcnn} are chosen as the student detector. The backbone of the student detector is ResNet50. Without distillation, Faster RCNN achieves 38.4\% mAP. The application of our DRKD brings an increase of 3.2\% mAP.
FGFI~\cite{fgfi} achieves 39.1\% mAP, which is 2.5\% mAP worse than our method. DRKD also exceeds GKD~\cite{Tang2022DistillingOD}, NLD~\cite{non_local_distill}.
Besides, Grid RCNN without distillation achieves 40.4\% mAP. NLD~\cite{non_local_distill} is a state-of-the-art distillation method, which gets 42.6\% mAP. Our DRKD obtains 43.0\%
mAP, exceeding NLD 0.4\% mAP. For Dynamic RCNN, DKRD achieves 43.0\% mAP which is higher than all other methods.

For the one-stage detector, RetinaNet with ResNeXt101 is selected as the teacher detector. RetinaNet~\cite{retinanet} with ResNet50 is chosen as the student detector. Without distillation, RetinaNet achieves 37.4\% mAP. Applying our DRKD brings an increase of 2.9\% mAP, achieving 40.3\% mAP. The mAP of DRKD is higher than that of FRS and NLD. Besides, RepPoints~\cite{reppoints} without distillation achieves 38.6\% mAP. Applying our DRKD brings an increase of 3.1\% mAP. The accuracy of the student detector even exceeds the teacher detector by 0.7\% mAP. It also exceeds FGD by 0.4\% mAP. The above results show that our distillation method achieves state-of-the-art performance on both one-stage and two-stage detectors.

\section{Conclusion}
 In this paper, we propose dual relation knowledge distillation. In it, the pixel-wise relation distillation helps to address the feature imbalance issue. Also, the instance-wise relation distillation is proposed to get a richer representation for small instances. We observe that the small instance has richer relations with other size instances. The experiment proves that the detection accuracy of small objects can be improved significantly by using our DRKD. The proposed distillation method is general and can be easily applied for both one-stage and two-stage detectors.
 %The experiments prove that our method is effective and achieves better performance, compared with state-of-the-art methods.

\section*{Contribution Statement}
Zhenliang Ni and Fukui Yang contribute equally to this paper, which is identified by * in the title.
\bibliographystyle{named}
\bibliography{eccv}

\begin{thebibliography}{}

\bibitem[\protect\citeauthoryear{Cai and Vasconcelos}{2018}]{cascade_rcnn}
Zhaowei Cai and Nuno Vasconcelos.
\newblock Cascade r-cnn: Delving into high quality object detection.
\newblock In {\em Proceedings of the IEEE conference on computer vision and
  pattern recognition}, pages 6154--6162, 2018.

\bibitem[\protect\citeauthoryear{Chen \bgroup \em et al.\egroup
  }{2017}]{chen2017learning}
Guobin Chen, Wongun Choi, Xiang Yu, Tony Han, and Manmohan Chandraker.
\newblock Learning efficient object detection models with knowledge
  distillation.
\newblock {\em Advances in neural information processing systems}, 30, 2017.

\bibitem[\protect\citeauthoryear{Chen \bgroup \em et al.\egroup }{2019}]{glore}
Yunpeng Chen, Marcus Rohrbach, Zhicheng Yan, Yan Shuicheng, Jiashi Feng, and
  Yannis Kalantidis.
\newblock Graph-based global reasoning networks.
\newblock In {\em Proceedings of the IEEE/CVF Conference on Computer Vision and
  Pattern Recognition}, pages 433--442, 2019.

\bibitem[\protect\citeauthoryear{Dai \bgroup \em et al.\egroup }{2021}]{gid}
Xing Dai, Zeren Jiang, Zhao Wu, Yiping Bao, Zhicheng Wang, Si~Liu, and Erjin
  Zhou.
\newblock General instance distillation for object detection.
\newblock In {\em Proceedings of the IEEE/CVF Conference on Computer Vision and
  Pattern Recognition}, pages 7842--7851, 2021.

\bibitem[\protect\citeauthoryear{Du \bgroup \em et al.\egroup }{2021}]{FRS}
Zhixing Du, Rui Zhang, Ming-Fang Chang, Xishan Zhang, Shaoli Liu, Tianshi Chen,
  and Yunji Chen.
\newblock Distilling object detectors with feature richness.
\newblock In {\em Neural Information Processing Systems}, 2021.

\bibitem[\protect\citeauthoryear{Duan \bgroup \em et al.\egroup
  }{2019}]{centernet}
Kaiwen Duan, Song Bai, Lingxi Xie, Honggang Qi, Qingming Huang, and Qi~Tian.
\newblock Centernet: Keypoint triplets for object detection.
\newblock In {\em Proceedings of the IEEE/CVF International Conference on
  Computer Vision}, pages 6569--6578, 2019.

\bibitem[\protect\citeauthoryear{Fu \bgroup \em et al.\egroup }{2019}]{dan}
Jun Fu, Jing Liu, Haijie Tian, Yong Li, Yongjun Bao, Zhiwei Fang, and Hanqing
  Lu.
\newblock Dual attention network for scene segmentation.
\newblock In {\em Proceedings of the IEEE/CVF Conference on Computer Vision and
  Pattern Recognition (CVPR)}, June 2019.

\bibitem[\protect\citeauthoryear{Gong \bgroup \em et al.\egroup
  }{2019}]{gong2019differentiable}
Ruihao Gong, Xianglong Liu, Shenghu Jiang, Tianxiang Li, Peng Hu, Jiazhen Lin,
  Fengwei Yu, and Junjie Yan.
\newblock Differentiable soft quantization: Bridging full-precision and low-bit
  neural networks.
\newblock In {\em Proceedings of the IEEE/CVF International Conference on
  Computer Vision}, pages 4852--4861, 2019.

\bibitem[\protect\citeauthoryear{Guo \bgroup \em et al.\egroup }{2021}]{defeat}
Jianyuan Guo, Kai Han, Yunhe Wang, Han Wu, Xinghao Chen, Chunjing Xu, and Chang
  Xu.
\newblock Distilling object detectors via decoupled features.
\newblock In {\em Proceedings of the IEEE/CVF Conference on Computer Vision and
  Pattern Recognition}, pages 2154--2164, 2021.

\bibitem[\protect\citeauthoryear{Han \bgroup \em et al.\egroup
  }{2015}]{han2015learning}
Song Han, Jeff Pool, John Tran, and William~J Dally.
\newblock Learning both weights and connections for efficient neural network.
\newblock In {\em NIPS}, 2015.

\bibitem[\protect\citeauthoryear{He \bgroup \em et al.\egroup }{2016}]{resnet}
Kaiming He, Xiangyu Zhang, Shaoqing Ren, and Jian Sun.
\newblock Deep residual learning for image recognition.
\newblock In {\em Proceedings of the IEEE Conference on Computer Vision and
  Pattern Recognition (CVPR)}, June 2016.

\bibitem[\protect\citeauthoryear{Heo \bgroup \em et al.\egroup
  }{2019}]{heo2019comprehensive}
Byeongho Heo, Jeesoo Kim, Sangdoo Yun, Hyojin Park, Nojun Kwak, and Jin~Young
  Choi.
\newblock A comprehensive overhaul of feature distillation.
\newblock In {\em Proceedings of the IEEE/CVF International Conference on
  Computer Vision}, pages 1921--1930, 2019.

\bibitem[\protect\citeauthoryear{Law and Deng}{2018}]{cornernet}
Hei Law and Jia Deng.
\newblock Cornernet: Detecting objects as paired keypoints.
\newblock In {\em Proceedings of the European conference on computer vision
  (ECCV)}, pages 734--750, 2018.

\bibitem[\protect\citeauthoryear{Lin \bgroup \em et al.\egroup }{2014}]{coco}
Tsung-Yi Lin, Michael Maire, Serge Belongie, James Hays, Pietro Perona, Deva
  Ramanan, Piotr Doll{\'a}r, and C~Lawrence Zitnick.
\newblock Microsoft coco: Common objects in context.
\newblock In {\em European conference on computer vision}, pages 740--755.
  Springer, 2014.

\bibitem[\protect\citeauthoryear{Lin \bgroup \em et al.\egroup
  }{2017}]{retinanet}
Tsung-Yi Lin, Priya Goyal, Ross Girshick, Kaiming He, and Piotr Doll{\'a}r.
\newblock Focal loss for dense object detection.
\newblock In {\em Proceedings of the IEEE international conference on computer
  vision}, pages 2980--2988, 2017.

\bibitem[\protect\citeauthoryear{Lin \bgroup \em et al.\egroup
  }{2020}]{lin2020hrank}
Mingbao Lin, Rongrong Ji, Yan Wang, Yichen Zhang, Baochang Zhang, Yonghong
  Tian, and Ling Shao.
\newblock Hrank: Filter pruning using high-rank feature map.
\newblock In {\em Proceedings of the IEEE/CVF Conference on Computer Vision and
  Pattern Recognition}, pages 1529--1538, 2020.

\bibitem[\protect\citeauthoryear{Liu \bgroup \em et al.\egroup }{2016}]{ssd}
Wei Liu, Dragomir Anguelov, Dumitru Erhan, Christian Szegedy, Scott Reed,
  Cheng-Yang Fu, and Alexander~C Berg.
\newblock Ssd: Single shot multibox detector.
\newblock In {\em European conference on computer vision}, pages 21--37.
  Springer, 2016.

\bibitem[\protect\citeauthoryear{Lu \bgroup \em et al.\egroup
  }{2019}]{grid_rcnn}
Xin Lu, Buyu Li, Yuxin Yue, Quanquan Li, and Junjie Yan.
\newblock Grid r-cnn.
\newblock In {\em Proceedings of the IEEE/CVF Conference on Computer Vision and
  Pattern Recognition}, pages 7363--7372, 2019.

\bibitem[\protect\citeauthoryear{Redmon and Farhadi}{2018}]{yolov3}
Joseph Redmon and Ali Farhadi.
\newblock Yolov3: An incremental improvement.
\newblock {\em arXiv preprint arXiv:1804.02767}, 2018.

\bibitem[\protect\citeauthoryear{Ren \bgroup \em et al.\egroup
  }{2015}]{faster_rcnn}
Shaoqing Ren, Kaiming He, Ross Girshick, and Jian Sun.
\newblock Faster r-cnn: Towards real-time object detection with region proposal
  networks.
\newblock {\em Advances in neural information processing systems}, 28:91--99,
  2015.

\bibitem[\protect\citeauthoryear{Sun \bgroup \em et al.\egroup }{2020}]{tar}
Ruoyu Sun, Fuhui Tang, Xiaopeng Zhang, Hongkai Xiong, and Qi~Tian.
\newblock Distilling object detectors with task adaptive regularization.
\newblock {\em arXiv preprint arXiv:2006.13108}, 2020.

\bibitem[\protect\citeauthoryear{Tang \bgroup \em et al.\egroup
  }{2022}]{Tang2022DistillingOD}
Sanli Tang, Zhongyu Zhang, Zhanzhan Cheng, Jing Lu, Yunlu Xu, Yi~Niu, and Fan
  He.
\newblock Distilling object detectors with global knowledge.
\newblock In {\em European Conference on Computer Vision}, 2022.

\bibitem[\protect\citeauthoryear{Tian \bgroup \em et al.\egroup }{2019}]{fcos}
Zhi Tian, Chunhua Shen, Hao Chen, and Tong He.
\newblock Fcos: Fully convolutional one-stage object detection.
\newblock In {\em Proceedings of the IEEE/CVF international conference on
  computer vision}, pages 9627--9636, 2019.

\bibitem[\protect\citeauthoryear{Wang \bgroup \em et al.\egroup
  }{2018}]{non_local}
Xiaolong Wang, Ross Girshick, Abhinav Gupta, and Kaiming He.
\newblock Non-local neural networks.
\newblock In {\em Proceedings of the IEEE conference on computer vision and
  pattern recognition}, pages 7794--7803, 2018.

\bibitem[\protect\citeauthoryear{Wang \bgroup \em et al.\egroup }{2019}]{fgfi}
Tao Wang, Li~Yuan, Xiaopeng Zhang, and Jiashi Feng.
\newblock Distilling object detectors with fine-grained feature imitation.
\newblock In {\em Proceedings of the IEEE/CVF Conference on Computer Vision and
  Pattern Recognition}, pages 4933--4942, 2019.

\bibitem[\protect\citeauthoryear{Xie \bgroup \em et al.\egroup
  }{2017}]{resnext}
Saining Xie, Ross Girshick, Piotr Dollar, Zhuowen Tu, and Kaiming He.
\newblock Aggregated residual transformations for deep neural networks.
\newblock In {\em Proceedings of the IEEE Conference on Computer Vision and
  Pattern Recognition (CVPR)}, July 2017.

\bibitem[\protect\citeauthoryear{Yang \bgroup \em et al.\egroup
  }{2019}]{reppoints}
Ze~Yang, Shaohui Liu, Han Hu, Liwei Wang, and Stephen Lin.
\newblock Reppoints: Point set representation for object detection.
\newblock In {\em Proceedings of the IEEE/CVF International Conference on
  Computer Vision}, pages 9657--9666, 2019.

\bibitem[\protect\citeauthoryear{Yang \bgroup \em et al.\egroup }{2022}]{fgd}
Zhendong Yang, Zhe Li, Xiaohu Jiang, Yuan Gong, Zehuan Yuan, Danpei Zhao, and
  Chun Yuan.
\newblock Focal and global knowledge distillation for detectors.
\newblock {\em 2022 IEEE/CVF Conference on Computer Vision and Pattern
  Recognition (CVPR)}, pages 4643--4652, 2022.

\bibitem[\protect\citeauthoryear{Zhang and Ma}{2021}]{non_local_distill}
Linfeng Zhang and Kaisheng Ma.
\newblock Improve object detection with feature-based knowledge distillation:
  Towards accurate and efficient detectors.
\newblock In {\em International Conference on Learning Representations}, 2021.

\bibitem[\protect\citeauthoryear{Zhang \bgroup \em et al.\egroup
  }{2020}]{dynamic_rcnn}
Hongkai Zhang, Hong Chang, Bingpeng Ma, Naiyan Wang, and Xilin Chen.
\newblock Dynamic r-cnn: Towards high quality object detection via dynamic
  training.
\newblock In {\em European Conference on Computer Vision}, pages 260--275.
  Springer, 2020.

\bibitem[\protect\citeauthoryear{Zhang \bgroup \em et al.\egroup
  }{2021}]{zhang2021diversifying}
Xiangguo Zhang, Haotong Qin, Yifu Ding, Ruihao Gong, Qinghua Yan, Renshuai Tao,
  Yuhang Li, Fengwei Yu, and Xianglong Liu.
\newblock Diversifying sample generation for accurate data-free quantization.
\newblock In {\em Proceedings of the IEEE/CVF Conference on Computer Vision and
  Pattern Recognition}, pages 15658--15667, 2021.

\end{thebibliography}

\end{document}